\documentclass{article} % For LaTeX2e
\usepackage{iclr2021,times}
\iclrfinalcopy

\usepackage{amsmath,amsfonts,bm}

\def\eqref#1{equation~\ref{#1}}
\def\1{\bm{1}}

\DeclareMathAlphabet{\mathsfit}{\encodingdefault}{\sfdefault}{m}{sl}
\SetMathAlphabet{\mathsfit}{bold}{\encodingdefault}{\sfdefault}{bx}{n}

\newcommand{\E}{\mathbb{E}}

\newcommand{\R}{\mathbb{R}}

\usepackage{enumitem}
\usepackage{hyperref}
\usepackage{url}
\usepackage{graphicx}
\usepackage{subcaption}

\title{Fourier Neural Operator for \\
Parametric Partial Differential Equations} 

\author{Zongyi Li \\zongyili@caltech.edu
\And Nikola Kovachki  \\nkovachki@caltech.edu
\And  Kamyar Azizzadenesheli \\kamyar@purdue.edu
\And Burigede Liu \\bgl@caltech.edu
\And Kaushik Bhattacharya \\bhatta@caltech.edu
\And  Andrew Stuart \\astuart@caltech.edu
\And Anima Anandkumar \\anima@caltech.edu
}

\newcommand{\A}{\mathcal{A}}
\newcommand{\C}{\mathbb{C}}
\newcommand{\U}{\mathcal{U}}

\newcommand{\Ftrue}{G^\dagger}
\newcommand{\F}{{G}}

\newcommand{\cG}{\mathcal{F}}

\newcommand{\cK}{\mathcal{K}}

\newcommand{\bbZ}{\mathbb{Z}}

\newtheorem{definition}{Definition}

\begin{document}

\maketitle

\begin{abstract}
The classical development of neural networks has primarily focused on learning mappings between finite-dimensional Euclidean spaces.  Recently, this has been generalized to neural operators that learn mappings between function spaces. For partial differential equations (PDEs), neural operators directly learn the mapping from any functional parametric dependence to the solution. Thus, they learn an entire family of PDEs, in contrast to classical methods which solve one instance of the equation. 
In this work, we formulate a new neural operator by parameterizing the integral kernel directly in Fourier space, allowing for an expressive and efficient architecture. 
We perform experiments on Burgers' equation, Darcy flow, and Navier-Stokes equation. The Fourier neural operator is the first ML-based method to successfully model turbulent flows with zero-shot super-resolution. It is up to three orders of magnitude faster compared to traditional PDE solvers. Additionally, it achieves superior accuracy compared to previous learning-based solvers under fixed resolution. 
\end{abstract}

\section{Introduction}
\label{sec:intro}
Many problems in science and engineering involve solving complex partial differential equation (PDE) systems repeatedly for different values of some parameters. Examples arise in molecular dynamics, micro-mechanics, and turbulent flows. Often such systems require fine discretization in order to capture the phenomenon being modeled. 
As a consequence, traditional numerical solvers are slow and sometimes inefficient. For example, when designing materials such as airfoils, one needs to solve the associated inverse problem where thousands of evaluations of the forward model are needed. A fast method can make such problems feasible. 

\paragraph{Conventional solvers vs. Data-driven methods.} 
Traditional solvers such as finite element methods (FEM) and finite difference methods (FDM) solve the equation by discretizing the space. Therefore, they impose a trade-off on the resolution: coarse grids are fast but less accurate; fine grids are accurate but slow. Complex PDE systems, as described above, usually require a very fine discretization, and therefore very challenging and time-consuming for traditional solvers. 
On the other hand, data-driven methods can directly learn the trajectory of the family of equations from the data. As a result, the learning-based method can be orders of magnitude faster than the conventional solvers. 

Machine learning methods may hold the key to revolutionizing scientific disciplines by providing fast solvers that approximate or enhance traditional ones \citep{raissi2019physics, jiang2020meshfreeflownet, greenfeld2019learning, kochkov2021machine}. However, classical neural networks map between finite-dimensional spaces and can therefore only learn solutions tied to a specific discretization. This is often a limitation for practical applications and therefore the development of mesh-invariant neural networks is required. 
We first outline two mainstream neural network-based approaches for PDEs -- the finite-dimensional operators and Neural-FEM.

\paragraph{Finite-dimensional operators.}
These approaches parameterize the solution operator as a deep convolutional neural network between finite-dimensional Euclidean spaces \cite{guo2016convolutional, Zabaras, Adler2017, bhatnagar2019prediction,khoo2017solving}.
Such approaches are, by definition, mesh-dependent and will need modifications and tuning for different resolutions and discretizations in order to achieve consistent error (if at all possible). Furthermore, these approaches are limited to the discretization size and geometry of the training data and hence, it is not possible to query solutions at new points in the domain. In contrast, we show, for our method, both invariance of the error to grid resolution, and the ability to transfer the solution between meshes.

\paragraph{Neural-FEM.}
The second approach directly parameterizes the solution function as a neural network 
\citep{Weinan, raissi2019physics,bar2019unsupervised,smith2020eikonet, pan2020physics}. This approach is designed to model one specific instance of the PDE, not the solution operator. It is mesh-independent and accurate, but for any given new instance of the functional parameter/coefficient, it requires training a new neural network. The approach closely resembles classical methods such as finite elements, replacing the linear span of a finite set of local basis functions with the space of neural networks.
%, but, as such, it also suffers from the same computational issues.
The Neural-FEM approach suffers from the same computational issue as classical methods: the optimization problem needs to be solved for every new instance. 
Furthermore, the approach is limited to a setting in which the underlying PDE is known. 
% Our methodology circumvents these issues.

\begin{figure}
    \centering
    \includegraphics[width=12cm]{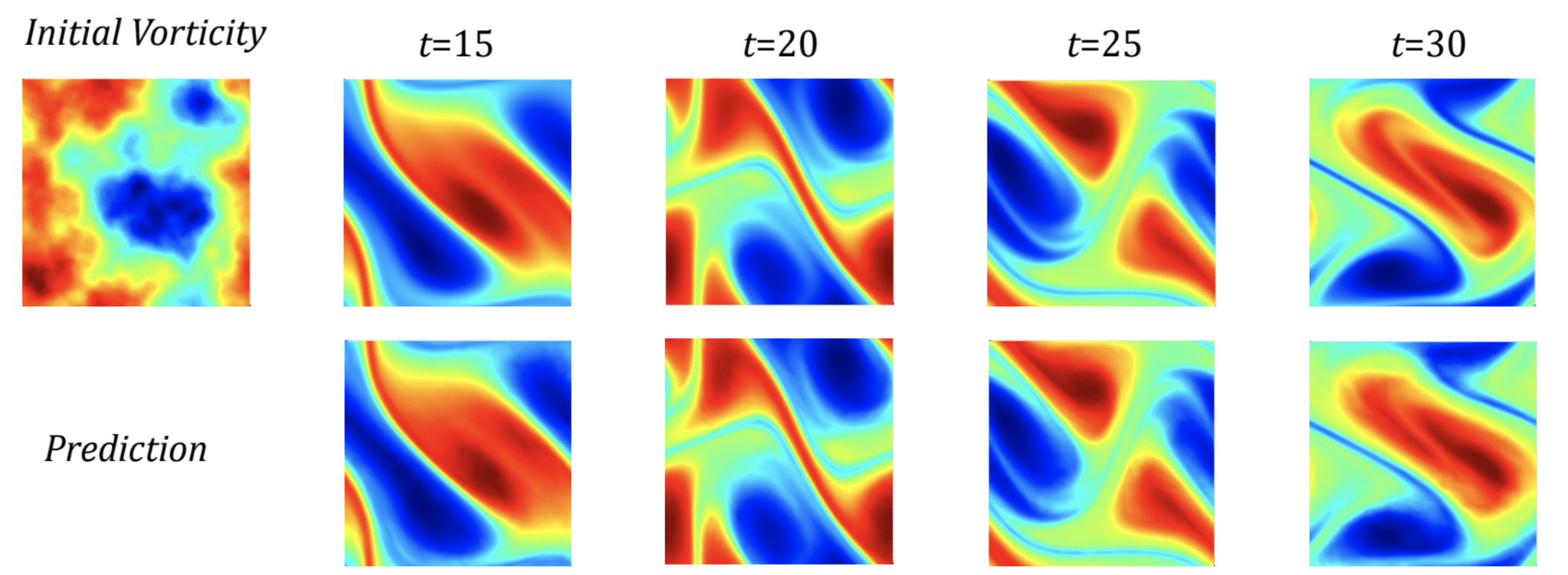}\\
    \small{
    % {\bf (a)} Start from input $v$. On top: apply the Fourier transform $\cG$; a linear transform $R$ on the lower Fourier modes and filters out the higher modes; then apply the inverse Fourier transform $\cG^{-1}$. On bottom: apply a local linear transform $W$. 
    Zero-shot super-resolution: Navier-Stokes Equation with viscosity $\nu=1\mathrm{e}{-4}$; Ground truth on top and prediction on bottom; trained on $64\times64\times20$ dataset; evaluated on $256\times256\times80$  (see Section \ref{sec:superresolution}).}
    \caption{ {\bf top:} The architecture of the Fourier layer; \textbf{bottom:} Example flow from Navier-Stokes.}
    \label{fig:1}
\end{figure}

\paragraph{Neural Operators.}
Recently, a new line of work proposed learning mesh-free, infinite-dimensional operators with neural networks
\citep{lu2019deeponet,Kovachki,nelsen2020random,li2020neural,li2020multipole, patel2021physics}.
The neural operator remedies the mesh-dependent nature of the finite-dimensional operator methods discussed above by producing a single set of network parameters that may be used with different discretizations. It has the ability to transfer solutions between meshes. 
Furthermore, the neural operator needs to be trained only once. Obtaining a solution for a new instance of the parameter requires only a forward pass of the network, alleviating the major computational issues incurred in Neural-FEM methods. 
Lastly, the neural operator requires no knowledge of the underlying PDE, only data. Thus far, neural operators have not yielded efficient numerical algorithms that can parallel the success of convolutional or recurrent neural networks in the finite-dimensional setting due to the cost of evaluating integral operators. Through the fast Fourier transform, our work alleviates this issue. 

%in the setting of mappings between function spaces, due to the high computational complexity of the integral operator. We aim to improve this with Fourier methods.

% \paragraph{Spectral Methods.}
% Spectral methods are techniques for solving differential equations by projecting onto a finite set of basis functions. When the basis are trigonometric, one may use the Fourier Transform by exploiting the fact that differentiation is multiplication in the Fourier domain. 
\paragraph{Fourier Transform.}
The Fourier transform is frequently used in spectral methods for solving differential equations, since differentiation is equivalent to multiplication in the Fourier domain.
%When the equation is smooth and well-conditioned, spectral methods usually have exponential convergence on error rate. 
Fourier transforms have also played an important role in the development of deep learning. In theory, they appear in the proof of the universal approximation theorem \citep{hornik1989multilayer} and, empirically, they have been used to speed up convolutional neural networks \citep{mathieu2013fast}. 
%They are used in the proof of the universal approximation theorem \citep{hornik1989multilayer} and to speed up convolution neural networks \citep{mathieu2013fast}. 
Neural network architectures involving the Fourier transform or the use of sinusoidal activation functions have also been proposed and studied \citep{bengio2007scaling,mingo2004Fourier, sitzmann2020implicit}.
Recently, some spectral methods for PDEs have been extended to neural networks \citep{fan2019bcr, fan2019multiscale, kashinath2020enforcing}. We build on these works by proposing a neural operator architecture defined directly in Fourier space with quasi-linear time complexity and state-of-the-art approximation capabilities. 
%Bearing these potential advantages of spectral methods, we propose a neural operator architecture based on convolution via Fourier transform which has quasi-linear time complexity and excellent accuracy rate.

\paragraph{Our Contributions.}
\label{ssec:OC}
We introduce the Fourier neural operator, a novel deep learning architecture able to learn mappings between infinite-dimensional spaces of functions; the integral operator is restricted to a convolution, and instantiated through a linear transformation in the Fourier domain.
\begin{itemize}[leftmargin=*]
\item The Fourier neural operator is the first work that learns the resolution-invariant solution operator for the family of Navier-Stokes equation in the turbulent regime, where previous graph-based neural operators do not converge.
\item By construction, the method shares the same learned network parameters irrespective of the discretization used on the input and output spaces. It can do zero-shot super-resolution: trained on a lower resolution directly evaluated on a higher resolution, as shown in Figure \ref{fig:1}.
\item The proposed method consistently outperforms all existing deep learning methods even when fixing the resolution to be $64 \times 64$. It achieves error rates that are $30\%$ lower on Burgers' Equation, $60\%$ lower on Darcy Flow, and $30\%$ lower on Navier Stokes (turbulent regime with viscosity $\nu=1\mathrm{e}{-4}$).
When learning the mapping for the entire time series, the method achieves $<1\%$ error with viscosity $\nu=1\mathrm{e}{-3}$ and $8\%$ error with viscosity $\nu=1\mathrm{e}{-4}$.
\item On a $256 \times 256$ grid, the Fourier neural operator has an inference time of only $0.005$s compared to the $2.2s$ of the pseudo-spectral method used to solve Navier-Stokes.
Despite its tremendous speed advantage, the method does not suffer from accuracy degradation when used in downstream applications such as solving the Bayesian inverse problem, as shown in Figure \ref{fig:baysian}.
%Despite the sharp speed advantage, our method can recover equally good initial conditions on the Bayesian inverse problem as shown in Figure \ref{fig:baysian}.
\end{itemize}
% We observe that the Fourier neural operator captures global interactions through convolution with low-frequency functions and returns high-frequency modes by composition with an activation function, allowing it to approximate functions with slow Fourier mode decay (Section \ref{sec:numerics}). Furthermore, local neural networks fix the periodic boundary which comes from the inverse Fourier transform and allows the method to approximate functions with any boundary conditions. We demonstrate this by having non-periodic boundaries on the spatial domain of Darcy flow and the time domain of the Navier-Stokes equation.
We observed that the proposed framework can approximate complex operators raising in PDEs that are highly non-linear, with high frequency modes and slow energy decay.
The power of neural operators comes from combining linear, global integral operators (via the Fourier transform) and non-linear, local activation functions.  Similar to the way standard neural networks approximate highly non-linear functions by combining linear multiplications with non-linear activations, the proposed neural operators can approximate highly non-linear operators.
%The Fourier neural operator has excellent accuracy because of the exponential decay of the spectrum and exceptional time complexity due to fast Fourier transform. While the Fourier layer truncates out the higher frequency modes, the Fourier neural operator as a whole can still approximate equations with non-trivial higher modes and slow spectral decay. We show in a spectral analysis that the Fourier neural operator can surpass spectral decomposition with the same truncation.

\section{Learning Operators}
\label{sec:operator}
Our methodology learns a mapping between two infinite dimensional spaces from a finite
collection of observed input-output pairs. Let  $D \subset \R^d$ be a bounded, open set and \(\A = \A(D;\R^{d_a})\) and \(\U= \U(D;\R^{d_u})\) be separable Banach spaces of function taking values in \(\R^{d_a}\) and \(\R^{d_u}\) respectively. Furthermore let \(\Ftrue : \A \to \U\) be a (typically) non-linear map. We study maps \(\Ftrue\) which arise as the solution operators of parametric PDEs -- see Section \ref{sec:numerics} for examples. Suppose we have observations \(\{a_j, u_j\}_{j=1}^N\) where 
\(a_j \sim \mu\) is an i.i.d. sequence from the probability measure \(\mu\) supported on 
\(\A\) and \(u_j = \Ftrue(a_j)\) is possibly corrupted with noise. We aim to build an approximation of \(\Ftrue\) by 
constructing a parametric map 
\begin{equation}
\label{eq:approxmap}
\F : \A \times \Theta \to \U
\qquad
\text{or equivalently,}
\qquad
\F_{\theta} : \A \to \U, \quad \theta \in \Theta
\end{equation}
for some finite-dimensional parameter space \(\Theta\) by choosing
\(\theta^\dagger \in \Theta\) so that \(\F(\cdot, \theta^\dagger) = \F_{\theta^\dagger} \approx \Ftrue\).
This is a natural framework for learning in infinite-dimensions as one could define a cost functional \(C : \U \times \U \to \R\) and seek a minimizer of the problem
\[\min_{\theta \in \Theta} \E_{a \sim \mu} [C(\F(a,\theta), \Ftrue(a))]\]
which directly parallels the classical finite-dimensional 
setting \citep{Vapnik1998}. Showing the existence of minimizers, in the infinite-dimensional setting, remains a challenging open problem. We will approach this problem in the test-train setting by using a data-driven
empirical approximation to the cost used to determine
$\theta$ and to test the accuracy of the approximation.
Because we conceptualize our methodology in the infinite-dimensional setting, all finite-dimensional approximations share a common set of parameters which are consistent in infinite dimensions. 
% Practically this means we obtain an approximation error that is independent of the function discretization -- a feature not shared by standard CNNs (see Figure \ref{fig:error}).
% To be concrete we will consider  Banach spaces of real-valued functions defined on a bounded open set $D$ in $\mathbb{R}^d$. We then consider
% mappings $\Ftrue$ which take input functions defined on $D$ and map them to another set of functions, also defined on $D$.
A table of notation is shown in Appendix \ref{table:burgers}.

\paragraph{Learning the Operator.} Approximating the operator \(\Ftrue\) is a different and typically much more challenging task than finding the solution \(u \in \U\) of a PDE for a single instance of the parameter \(a \in \A\). Most existing methods, ranging from classical finite elements, finite differences, and finite volumes to modern machine learning approaches such as physics-informed neural networks (PINNs) \citep{raissi2019physics} aim at the latter and can therefore be computationally expensive. This makes them impractical for applications where a solution to the PDE is required for many different instances of the parameter. On the other hand, our approach directly approximates the operator and is therefore much cheaper and faster, offering tremendous computational savings when compared to traditional solvers. For an example application to Bayesian inverse problems, see Section \ref{sec:bayesian}.

%We want to stress that learning the operator $\F$ is more challenging and ambitious compared to finding the solution $u$ for a single instant $(a,u)$. Most of the existing methods, ranging from the traditional finite element, finite difference methods, to machine learning based physics-informed neural networks (PINN) \citep{raissi2019physics}, all aim to find $u$ for a single instant. On the contrary, we want to learn the operator $\F$ as a mapping from $a_j$ to $u_j$.
%Compared to PDE solvers such as FEM and PINN, the neural operator doesn't require to solve each equation. It can immediately output the evaluation for any new query of $a$. It is much faster when multiple evaluations are needed. For example, for the inverse problem when one needs to find some optimal material structures $a^*$, the neural operator can quickly evaluate any guesses $a'$, combining with Bayesian methods or optimization methods to find the optimal initial condition $a' \to a^*$, an example is shown in Section \ref{sec:bayesian}.

\paragraph{Discretization.} Since our data \(a_j\) and \(u_j\) are, in general, functions, to work with them numerically, we assume access only to point-wise evaluations. 
Let \(D_j = \{x_1,\dots,x_n\} \subset D\) be a \(n\)-point discretization of the domain \(D\) and assume we have observations \(a_j|_{D_j} \in \R^{n \times d_a}\), \(u_j|_{D_j} \in \R^{n\times d_v}\), for a finite collection  of input-output pairs indexed by $j$.
To be discretization-invariant, the neural operator can produce an answer \(u(x)\) for any \(x \in D\), potentially $x \notin D_j$.
Such a property is highly desirable as it allows a transfer of solutions between different grid geometries and discretizations.
% \kamyar{you might want to cite meshfreeflownet here}
% Zongyi: for this version we only have 8 pages, let's cite it in the camera ready version where we get one more page

%We note that, while the application of our methodology is based on having point-wise evaluations of the function, it is not limited by it. One may, for example, represent a function numerically as a finite set of truncated basis coefficients. The invariance of the representation would then be with respect to the size of this set. Our methodology can, in principle, be modified to accommodate this scenario through a suitably chosen architecture. We do not pursue this direction in the current work.

\begin{figure}
    \centering
    \includegraphics[width=14cm]{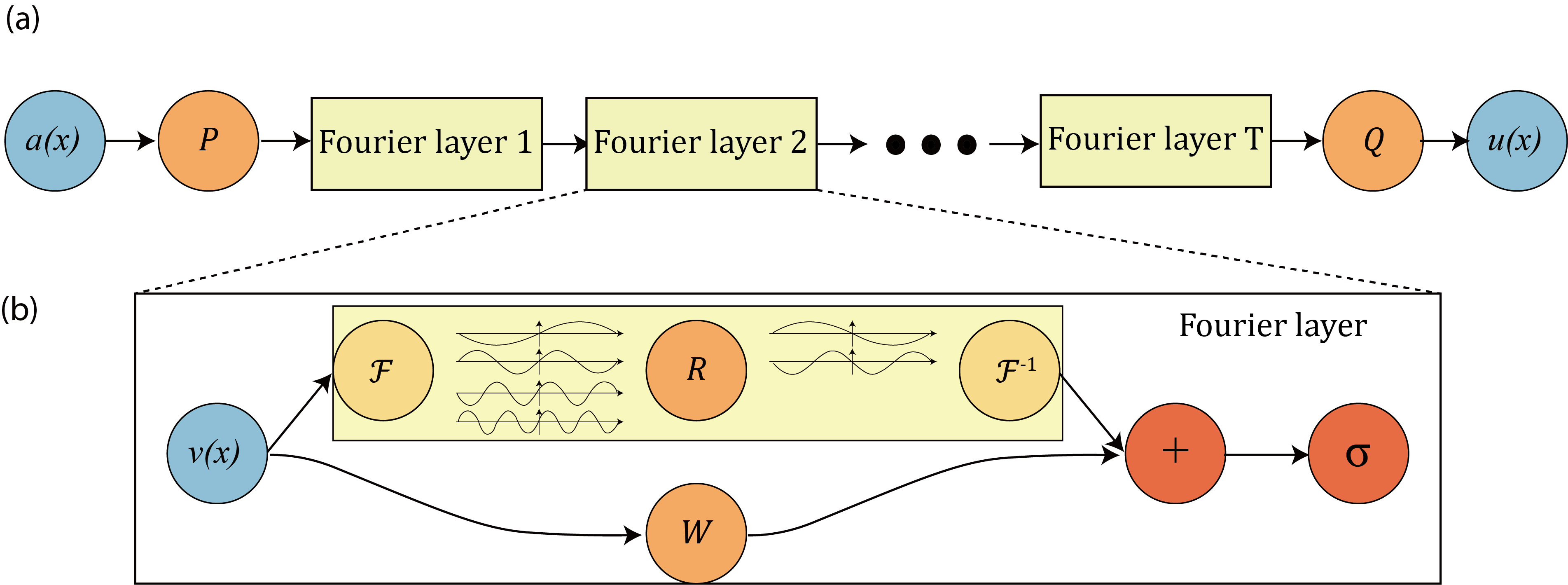}\\
    \small{
    {\bf (a) The full architecture of neural operator}: start from input $a$. 1. Lift to a higher dimension channel space by a neural network $P$. 2. Apply four layers of integral operators and activation functions. 3. Project back to the target dimension by a neural network $Q$. Output $u$.
    {\bf (b) Fourier layers}: Start from input $v$. On top: apply the Fourier transform $\cG$; a linear transform $R$ on the lower Fourier modes and filters out the higher modes; then apply the inverse Fourier transform $\cG^{-1}$. On the bottom: apply a local linear transform $W$.
    }
    \caption{ {\bf top:} The architecture of the neural operators; \textbf{bottom:} Fourier layer.}
    \label{fig:arch}
\end{figure}

\section{Neural Operator}
The neural operator, proposed in \citep{li2020neural}, is formulated as an iterative architecture $v_0 \mapsto v_1 \mapsto \ldots \mapsto v_T$ where $v_j$ for $j=0,1,\dots,T-1$
is a sequence of functions each taking values in $\R^{d_v}$. As shown in Figure \ref{fig:arch} (a), the input \(a \in \A\) is first lifted to a higher dimensional representation $v_0(x) = P(a(x))$ by the local transformation \(P\) which is usually parameterized by a shallow fully-connected neural network. Then we apply several iterations of updates $v_t \mapsto v_{t+1}$ (defined below). The output $u(x) = Q(v_T(x))$ is the projection of $v_T$ by the local transformation $Q: \R^{d_v} \to \R^{d_u}$. In each iteration, the update $v_t \mapsto v_{t+1}$ is defined as the composition of a non-local integral operator $\cK$ and a local, nonlinear activation function $\sigma$.

%for some \(d_v > d_a\) and \(j=0,1,\dots,T-1\). The input $v_0 = P(a)$ is lifted to the higher dimension \(d_v\) (channel) by the local transformation $P: $  and outputs $u = Q(v_T)$, where $P:\A \to \U^{d_v}, Q:\U^{d_v} \to \U$ are local transformations (usually shallow neural networks) that encode the input $a(x)$ from dimension $d$ to a higher (channel) dimension $d_v$ and that decode $v_T(x)$ from the channel dimension $d_v$ to $d$. In each iteration, the update $v_t \mapsto v_{t+1}$ is defined as the composition of a non-local integral operator $\cK$ and a local, nonlinear activation function $\sigma$. 
\begin{definition}[Iterative updates]
Define the update to the representation $v_t \mapsto v_{t+1}$ by
%For each iteration, we update the representation $v_t \mapsto v_{t+1}$ by applying integration operator $\cK$, linear transformation $W$, and activation function $\sigma$:
\begin{equation}\label{def:int}
v_{t+1}(x) := \sigma\Big( W v_t(x) 
+ \bigl(\cK(a;\phi)v_t\bigr)(x) \Big), \qquad \forall x \in D
\end{equation}
where $\cK: \A \times \Theta_{\cK} \to \mathcal{L}(\U(D; \R^{d_v}),\U(D; \R^{d_v}))$ maps to bounded linear operators on $\U(D; \R^{d_v})$ and is parameterized by $\phi \in \Theta_{\cK}$, $W: \R^{d_v} \to \R^{d_v}$ is a linear transformation, and $\sigma : \R \to \R$ is a non-linear activation function whose action is defined component-wise. 
\end{definition}
We choose $\cK(a;\phi)$ to be a kernel integral transformation parameterized by a neural network.
\begin{definition}[Kernel integral operator $\cK$] 
Define the kernel integral operator mapping in (\ref{def:int}) by
\begin{equation}
\label{def:K_int}
\bigl(\cK(a;\phi)v_t\bigr)(x) :=  
\int_{D} \kappa\big(x,y,a(x),a(y);\phi\big) v_t(y) \mathrm{d}y, \qquad \forall x \in D
\end{equation}
where $\kappa_{\phi}: \R^{2(d+d_a)} \to \R^{d_v \times d_v}$ is a neural network parameterized by $\phi \in \Theta_{\cK}$.
\end{definition}
Here $\kappa_\phi$ plays the role of a kernel function which we learn from data. Together definitions 1 and 2 constitute a generalization of neural networks to infinite-dimensional spaces as first proposed in \cite{li2020neural}. Notice even the integral operator is linear, the neural operator can learn highly non-linear operators by composing linear integral operators with non-linear activation functions, analogous to standard neural networks.

If we remove the dependence on the function \(a\) and impose  $\kappa_{\phi}(x,y) = \kappa_{\phi}(x-y)$, we obtain that (\ref{def:K_int}) is a convolution operator, which is a natural choice from the perspective of fundamental solutions. We exploit this fact in the following section by parameterizing $\kappa_{\phi}$ directly in Fourier space and using the Fast Fourier Transform (FFT) to efficiently compute (\ref{def:K_int}). This leads to a fast architecture that obtains state-of-the-art results for PDE problems.

%It can be viewed as the kernel function $\kappa_a(\cdot,\cdot): \R^2 \to \R^{d_v \times d_v}$ that output the transformation for each pair $(x,y)$. In particular, if $k(x,y) = k(x-y)$, the integration is a convolution.

\section{Fourier Neural Operator}
\label{sec:fourier}
We propose replacing the kernel integral operator in (\ref{def:K_int}), by a convolution operator defined in Fourier space.  Let \(\cG\) denote the Fourier transform of a function $f: D \to \R^{d_v}$ and $\cG^{-1}$ its inverse then
\begin{align*}
    (\cG f)_j(k) = \int_{D} f_j(x) e^{- 2i \pi \langle x, k \rangle} \mathrm{d}x, \qquad
    (\cG^{-1} f)_j(x) = \int_{D} f_j(k) e^{2i \pi \langle x, k \rangle} \mathrm{d}k
\end{align*}
%The Fourier method replaces the kernel integration by a convolution. Using $\cG$ to denote the Fourier transform from function of spatial locations $x \in D$ to function of Fourier modes $k \in \bbZ^d$, and $\cG^{-1}$ its inverse:
%\begin{align*}
%    (\cG f)(k) = \int_{D} f(x) e^{- 2i \pi \langle x, k \rangle} \mathrm{d}x \qquad
%    (\cG^{-1} g)(x) = \sum_{k\in \bbZ^d} g(k) e^{2i \pi \langle x, k \rangle} 
%\end{align*}
for $j=1,\dots,d_v$ where \(i = \sqrt{-1}\) is the imaginary unit. By letting $\kappa_{\phi}(x,y,a(x),a(y)) = \kappa_{\phi}(x-y)$ in (\ref{def:K_int}) and applying the convolution theorem, we find that
\[\bigl(\cK(a;\phi)v_t\bigr)(x) = \cG^{-1} \bigl( \cG(\kappa_\phi) \cdot \cG(v_t) \bigr )(x), \qquad \forall x \in D. \]
We, therefore, propose to directly parameterize $\kappa_\phi$ in Fourier space.
\begin{definition}[Fourier integral operator $\cK$] Define the Fourier integral operator
\begin{equation}
\label{eq:Fourier}
\bigl(\cK(\phi)v_t\bigr)(x)=   
\cG^{-1}\Bigl(R_\phi \cdot (\cG v_t) \Bigr)(x) \qquad \forall x \in D 
\end{equation}
where $R_\phi$ is the Fourier transform of a periodic function $\kappa: \bar{D} \to \R^{d_v \times d_v}$ parameterized by \(\phi \in \Theta_\cK\).
An illustration is given in Figure \ref{fig:arch} (b).
\end{definition}

For frequency mode \(k \in D\), we have $(\cG v_t)(k) \in \C^{d_v}$ and $R_\phi(k) \in \C^{d_v \times d_v}$. Notice that since we assume $\kappa$ is periodic, it admits a Fourier series expansion, so we may work with the discrete modes $k \in \mathbb{Z}^d$. We pick a finite-dimensional parameterization by truncating the Fourier series at a maximal number of modes 
\(k_{\text{max}} = |Z_{k_{\text{max}}}| = |\{k \in \mathbb{Z}^d : |k_j| \leq k_{\text{max},j}, \text{ for } j=1,\dots,d\}|.\)
We thus parameterize $R_\phi$ directly as complex-valued $(k_{\text{max}} \times d_v \times d_v)$-tensor comprising a collection of truncated Fourier modes and therefore drop $\phi$ from our notation. Since $\kappa$ is real-valued, we impose conjugate symmetry. 
% i.e.
% \[R(-k)_{j,l} = R^*(k)_{j,l} \qquad \forall k \in Z_{k_{\text{max}}}, \quad j,l=1,\dots,d_v.\]
We note that the set $Z_{k_{\text{max}}}$ is not the canonical choice for the low frequency modes of $v_t$. Indeed, the low frequency modes are usually defined by placing an upper-bound on the $\ell_1$-norm of $k \in \mathbb{Z}^d$. We choose $Z_{k_{\text{max}}}$ as above since it allows for an efficient implementation. 

%where the multiplication $\cdot$ is pointwise with respect to the mode $k$.  $(\cG v)(k) \in \C^{d_v}$ is the Fourier transform of $v$ on the mode $k$. $R(k) \in \C^{d_v \times d_v}$ is the weight matrix that is constrained so that the left hand-side of the identity in \eqref{eq:Fourier} is real-valued. We only pass in lower modes $k \leq k_{max}$; higher modes are filter out.
%In words, the Fourier integral operator consists of three steps 1. apply Fourier transformation  $v\mapsto \cG v$. 2. For each mode $k$ smaller than $k_{max}$, multiply $(\cG v)(k)$ with the weight $R(k)$.  Higher modes $(k > k_{max})$ are set to zeros. 3. in the end, apply the inverse Fourier transfer $\cG^{-1}$.
%The flow diagram is shown in Figure \ref{fig:1}.

\paragraph{The discrete case and the FFT.}
Assuming the domain $D$ is discretized with $n \in \mathbb{N}$ points, we have that $v_t \in \R^{n \times d_v}$ and $\cG (v_t) \in \C^{n \times d_v}$. Since we convolve $v_t$ with a function which only has $k_{\text{max}}$ Fourier modes, we may simply truncate the higher modes to obtain $\cG (v_t) \in \C^{k_{\text{max}} \times d_v}$. Multiplication by the weight tensor $R \in \C^{k_{\text{max}} \times d_v \times d_v}$ is then
%In the discretized case, if the discretization $D_j$ consists of $n$ points, then $v \in \R^{n \times d_v}$,  $k \in \{0,\ldots, n-1\}$, and $\cG v \in \C^{n \times d_v}$.  
%We truncate the modes up to $k_{max}$, so we can also write $\cG v \in \C^{k_{max} \times d_v}$.
%The weight matrix $R \in \C^{k_{max} \times d_v \times d_v}$ is a transformation on the channel dimension $d_v$, separately on each mode $k$:
\begin{equation}
\label{eq:fft_mult}
\bigl( R \cdot (\cG v_t) \bigr)_{k,l} = \sum_{j=1}^{d_v} R_{k,l,j}  (\cG v_t)_{k,j}, \qquad k=1,\dots,k_{\text{max}}, \quad j=1,\dots,d_v.
\end{equation}
%The transformed representation $R \cdot (\cG v)$ will have the same shape as the original one $\cG v \in \C^{n \times d_v}$.\\
When the discretization is uniform with resolution \(s_1 \times  \cdots \times s_d = n\), $\cG$ can be replaced by the Fast Fourier Transform. For $f \in \R^{n \times d_v}$,   $k = (k_1, \ldots, k_{d}) \in \bbZ_{s_1} \times \cdots \times \bbZ_{s_d}$, and $x=(x_1, \ldots, x_{d}) \in D$, the FFT $\hat{\cG}$ and its inverse $\hat{\cG}^{-1}$ are defined as
\begin{align*}
    (\hat{\cG} f)_l(k) = \sum_{x_1=0}^{s_1-1} \cdots \sum_{x_{d}=0}^{s_d-1} f_l(x_1, \ldots, x_{d}) e^{- 2i \pi \sum_{j=1}^{d} \frac{x_j k_j}{s_j} }, \\
    (\hat{\cG}^{-1} f)_l(x) = \sum_{k_1=0}^{s_1-1} \cdots \sum_{k_{d}=0}^{s_d-1} f_l(k_1, \ldots, k_{d}) e^{2i \pi \sum_{j=1}^{d} \frac{x_j k_j}{s_j} }
\end{align*}
for $l=1,\dots,d_v$. 
% where we abuse notation and index the rows of the matrix $f$ by either the Fourier mode $k$ or the spatial location $x$ and index the column by the subscript $l$.
In this case, the set of truncated modes becomes
\[Z_{k_{\text{max}}} = \{(k_1, \ldots, k_{d}) \in \bbZ_{s_1} \times \cdots \times \bbZ_{s_d} \mid k_j \leq k_{\text{max},j} \text{ or }\ s_j-k_j \leq k_{\text{max},j}, \text{ for } j=1,\dots,d\}.\]
When implemented, $R$ is treated as a $(s_1 \times \cdots \times s_d \times d_v \times d_v)$-tensor and the above definition of $Z_{k_{\text{max}}}$ corresponds to the ``corners'' of $R$, which allows for a straight-forward parallel implementation of (\ref{eq:fft_mult}) via matrix-vector multiplication. 
In practice, we have found that choosing $k_{\text{max},j} = 12$  which yields $k_{\text{max}} = 12^d$ parameters per channel to be sufficient for all the tasks that we consider.

%We note, however, that this is not the canonical choice for the low modes which are usually defined by
%\{\}

%lower modes are defined as $\{k = [k_1, \ldots, k_{d}] \mid k_i \leq k_{max,i}\ {\text or}\ s-k_i \leq k_{max,i}, \ {\text for} \ i=1,\ldots,d\}$, which are the ``corners'' of the $s \times \cdots \times s$ matrix (tensor) $Fv$, allowing easy parallel implementation via matrix-vector multiplication. In this case $k_{max} = 2^d \prod k_{max,i}$. Notice we don't use the canonical choice of lower mode ($\sum_i k_i \leq k_{max}$) for implementation reason. In practice, $k_{max,i} = 20,  (k_{max} = 20^d)$ is usually sufficient, no matter the given resolution $s$.

\paragraph{Parameterizations of $R$.} 
In general, $R$ can be defined to depend on $(\cG a)$ to parallel (\ref{def:K_int}).
Indeed, we can define $R_\phi: \mathbb{Z}^d \times \R^{d_v} \to \R^{d_v \times d_v}$
as a parametric function that maps \(\bigl(k,(\cG a)(k))\) to the values of the appropriate Fourier modes. We have experimented with linear as well as neural network parameterizations of $R_\phi$.
We find that the linear parameterization has a similar performance to the previously described direct parameterization,
% however it is not as efficient in both time complexity and the number of parameters required.
while neural networks have worse performance. This is likely due to the discrete structure of the space $\mathbb{Z}^d$.  
% Generally, we find that the influence of the direct dependence on $a$ is problem-dependent. Indeed, when $a$ is an initial condition, for example in the problems presented in Sections \ref{ssec:burgers} and \ref{sec:ns}, a direct dependence is unnecessary, while, when it is a geometric parameter such as in the problem presented in Section \ref{sec:darcy}, it may be beneficial.
Our experiments in this work focus on the direct parameterization presented above.

\paragraph{Invariance to discretization.}
The Fourier layers are discretization-invariant because they can learn from and evaluate functions which are discretized in an arbitrary way. Since parameters are learned directly in Fourier space, resolving the functions in physical space simply amounts to projecting on the basis $e^{2\pi i \langle x, k \rangle}$ which are well-defined everywhere on $\R^d$. This allows us to achieve zero-shot super-resolution as shown in Section \ref{sec:superresolution}.
% \nik{Therefore we achieve super-resolution (I'm not sure about this, how it should be stated and if we have an example?)}
% Zongyi: I guess we can leave the example to the experiment section
Furthermore, our architecture has a consistent error at any resolution of the inputs and outputs. On the other hand, notice that, in Figure \ref{fig:error}, the standard CNN methods we compare against have an error that grows with the resolution.  

%in a sense that it can learn and evaluate on arbitrary resolutions, because the modes $k_{max}$ is fixed and the corresponding basis function $e^{2i\pi \langle x,k\rangle}$ are resolution-invariant.
%When using the general Fourier transform, the Fourier neural operator is also discretization-invariant, while aliasing problem may raise. 
%When the input is uniform and FFT is used, the method will be restricted to uniform grids, but still resolution-invariant if one resolution is a multiple of another.

\paragraph{Quasi-linear complexity.}
The weight tensor $R$ contains $k_{\text{max}} < n$ modes, so the inner multiplication has complexity $O(k_{\text{max}})$. Therefore, the majority of the computational cost lies in computing the Fourier transform $\cG(v_t)$ and its inverse. General Fourier transforms have complexity $O(n^2)$, however, since we truncate the series the complexity is in fact $O(n k_{\text{max}})$, while the FFT has complexity $O(n \log n)$. Generally, we have found using FFTs to be very efficient. However a uniform discretization is required.

\section{Numerical experiments}
\label{sec:numerics}
In this section, we compare the proposed Fourier neural operator with multiple finite-dimensional architectures as well as operator-based approximation methods on the 1-d Burgers' equation, the 2-d Darcy Flow problem, and 2-d Navier-Stokes equation. The data generation processes are discussed in Appendices \ref{app:burgers},  \ref{app:darcy}, and \ref{app:ns} respectively. We do not compare against traditional solvers (FEM/FDM) or neural-FEM type methods since our goal is to produce an efficient operator approximation that can be used for downstream applications. We demonstrate one such application to the Bayesian inverse problem in Section \ref{sec:bayesian}.
%Notices we do not directly compare with traditional solvers (FEM/FDM) or neural-FEM type methods which only solve for one instance as discussed in section \ref{sec:intro}  and \ref{sec:operator}. Instead, we present a Bayesian inverse problem to demonstrate our speed advantage over the traditional solvers.

We construct our Fourier neural operator by stacking four Fourier integral operator layers as specified in (\ref{def:int}) and (\ref{eq:Fourier}) with the ReLU activation as well as batch normalization. Unless otherwise specified, we use $N=1000$ training instances and $200$ testing instances. We use Adam optimizer to train for $500$ epochs with an initial learning rate of $0.001$ that is halved every $100$ epochs. We set $k_{\text{max},j} = 16, d_v=64$ for the 1-d problem and $k_{\text{max},j} = 12, d_v=32$ for the 2-d problems. 
Lower resolution data are downsampled from higher resolution. 
All the computation is carried on a single Nvidia V100 GPU with 16GB memory.

\begin{figure}
    \centering
    \includegraphics[width=\textwidth]{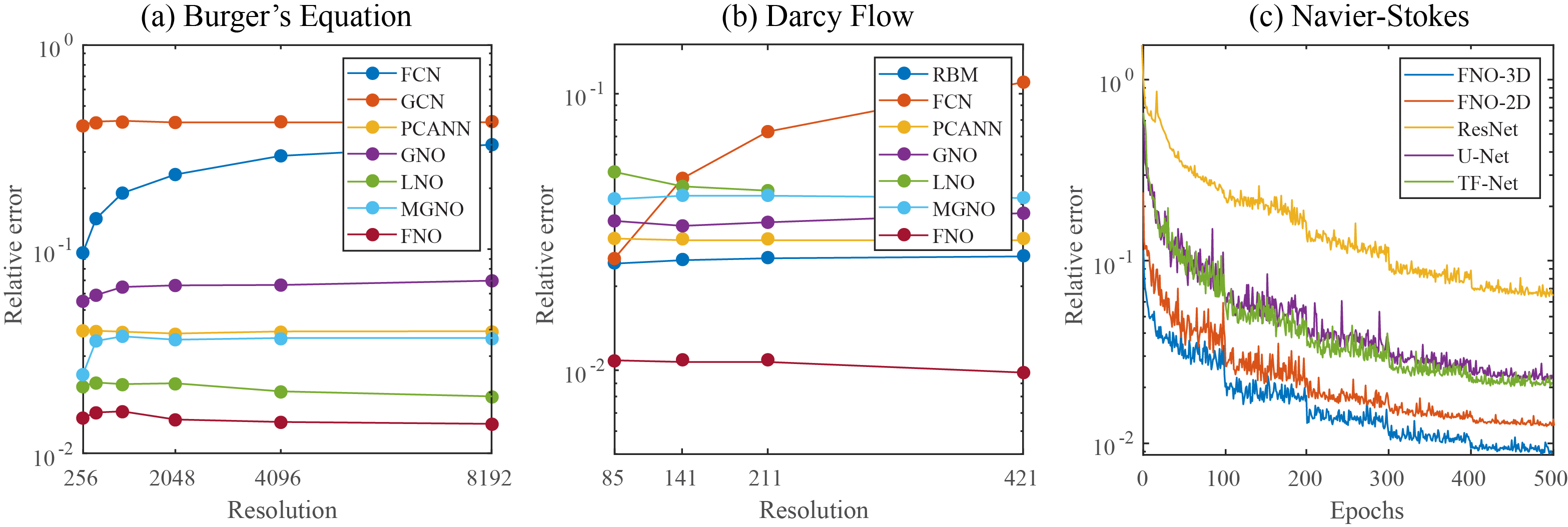}
    {\small \textbf{Left:} benchmarks on Burgers equation;  \textbf{Mid:} benchmarks on Darcy Flow for different resolutions; \textbf{Right:} the learning curves on Navier-Stokes $\nu=1\mathrm{e}{-3}$ with different benchmarks. Train and test on the same resolution.
    For acronyms, see Section \ref{sec:numerics}; details in Tables \ref{table:ns}, \ref{table:burgers}, \ref{table:darcy}.} 
    \caption{Benchmark on Burger's equation, Darcy Flow, and Navier-Stokes}
    \label{fig:error}
\end{figure}

\paragraph{Remark on Resolution.}
Traditional PDE solvers such as FEM and FDM approximate a single function and therefore their error to the continuum decreases as the resolution is increased. On the other hand, operator approximation is independent of the ways its data is discretized as long as all relevant information is resolved. Resolution-invariant operators have consistent error rates among different resolutions as shown in Figure \ref{fig:error}. Further, resolution-invariant operators can do zero-shot super-resolution, as shown in Section \ref{sec:superresolution}.

%Higher resolution shall lead to better accuracy for traditional methods (FEM/FDM) because they need to approximate the continuous function by discretization and the error accumulates through iterations, which is not the case for data-driven methods. The accuracy of data-driven methods depends primarily on the quality of the data. Assume a sufficiently large resolution ($s\geq32$), when the data are downsampled from the same ground truth, ideally data-driven should learn the operator with a similar accuracy rate.
%This is indeed the resolution-invariant property of neural operator-based methods. On the other hand, convolution neural networks based methods are usually tuned for a specific resolution and not resolution-invariant.

\paragraph{Benchmarks for time-independent problems (Burgers and Darcy):}
{\bf NN:} a simple point-wise feedforward neural network. 
{\bf RBM:} the classical Reduced Basis Method (using a POD basis) \citep{DeVoreReducedBasis}. 
{\bf FCN:} a the-state-of-the-art neural network architecture based on Fully Convolution Networks \citep{Zabaras}. 
% It has a good performance for small grids $s=61$. But fully convolution networks are mesh-dependent and therefore their error grows when the architecture moves to a larger grid.
{\bf PCANN:} an operator method using PCA as an autoencoder on both the input and output data and interpolating the latent spaces with a neural network \citep{Kovachki}.   
{\bf GNO:} the original graph neural operator \citep{li2020neural}.
{\bf MGNO:} the multipole graph neural operator \citep{li2020multipole}.
{\bf LNO:} a neural operator method based on the low-rank decomposition of the kernel $\kappa(x,y) := \sum^r_{j=1} \phi_j(x) \psi_j(y)$, similar to the unstacked DeepONet  proposed in \citep{lu2019deeponet}.
{\bf FNO:} the newly purposed Fourier neural operator. 

\paragraph{Benchmarks for time-dependent problems (Navier-Stokes):}
{\bf ResNet:} $18$ layers of 2-d convolution with residual connections \citep{he2016deep}.
{\bf U-Net:} A popular choice for image-to-image regression tasks consisting of four blocks with 2-d convolutions and deconvolutions \citep{ronneberger2015u}.
{\bf TF-Net:} A network designed for learning turbulent flows based on a combination of spatial and temporal convolutions \citep{wang2020towards}. 
{\bf FNO-2d:} 2-d Fourier neural operator with a RNN structure in time.
{\bf FNO-3d:} 3-d Fourier neural operator that directly convolves in space-time.

\subsection{Burgers' Equation}
\label{ssec:burgers}
The 1-d Burgers' equation is a non-linear PDE with various applications including modeling the one dimensional flow of a viscous fluid. It takes the form
\begin{align}
    \begin{split}
    \partial_t u(x,t) + \partial_x ( u^2(x,t)/2) &= \nu \partial_{xx} u(x,t), \qquad x \in (0,1), t \in (0,1] \\
    u(x,0) &= u_0(x), \qquad \qquad \:\: x \in (0,1)
    \end{split}
\end{align}
with periodic boundary conditions where $u_0 \in L^2_{\text{per}}((0,1);\R)$ is the initial condition and $\nu \in \R_+$ is the viscosity coefficient. We aim to learn the operator mapping the initial condition to the solution at time one, $\Ftrue: L^2_{\text{per}}((0,1);\R) \to H^r_{\text{per}} ((0,1);\R)$ defined by $u_0 \mapsto u(\cdot, 1)$ for any $r > 0$.
%We want to learn the operator from the initial condition to the final step: $\Ftrue: u(0) \mapsto u(1)$

The results of our experiments are shown in Figure \ref{fig:error} (a) and Table \ref{table:burgers} (Appendix \ref{app:burgers}). Our proposed method obtains the lowest relative error compared to any of the benchmarks. Further, the error is invariant with the resolution, while the error of convolution neural network based methods (FCN) grows with the resolution. 
Compared to other neural operator methods such as GNO and MGNO that use Nystr\"om sampling in physical space, the Fourier neural operator is both more accurate and more computationally efficient.

\subsection{Darcy Flow}
\label{sec:darcy}
We consider the steady-state of the 2-d Darcy Flow equation on the unit box which is the second order, linear, elliptic PDE
\begin{align}\label{ssec:darcy}
\begin{split}
- \nabla \cdot (a(x) \nabla u(x)) &= f(x) \qquad x \in (0,1)^2 \\
u(x) &= 0 \qquad \quad \:\:x \in \partial (0,1)^2
\end{split}
\end{align}
with a Dirichlet boundary where $a \in L^\infty((0,1)^2;\R_+)$  is the diffusion coefficient and $f \in L^2((0,1)^2;\R)$ is the forcing function. This PDE has numerous applications including modeling the pressure of subsurface flow, the deformation of linearly elastic materials, and the electric potential in conductive materials. We are interested in learning the operator mapping the diffusion coefficient to the solution, 
$\Ftrue: L^\infty((0,1)^2;\R_+) \to H^1_0 ((0,1)^2;\R_+)$ defined by $a \mapsto u$. Note that although the PDE is linear, the operator $\Ftrue$ is not.

%We want to learn the operator for the coefficient function $a$ to the solution function $u$:
%$\Ftrue: a \mapsto u$

The results of our experiments are shown in Figure \ref{fig:error} (b) and Table \ref{table:darcy} (Appendix \ref{app:darcy}). The proposed Fourier neural operator obtains nearly one order of magnitude lower relative error compared to any benchmarks. We again observe the invariance of the error with respect to the resolution. 

\subsection{ Navier-Stokes Equation}
\label{sec:ns}
We consider the 2-d Navier-Stokes equation for a viscous, incompressible fluid in vorticity form on the unit torus:
% {\bf equations under construction}\\
\begin{align}
\begin{split}
\partial_t w(x,t) + u(x,t) \cdot \nabla w(x,t) &= \nu \Delta w(x,t) + f(x), \qquad x \in (0,1)^2, t \in (0,T]  \\
\nabla \cdot u(x,t) &= 0, \qquad \qquad \qquad \qquad \quad x \in (0,1)^2, t \in [0,T]  \\
w(x,0) &= w_0(x), \qquad \qquad \qquad \quad x \in (0,1)^2 
\end{split}
\end{align}
where $u \in C([0,T]; H^r_{\text{per}}((0,1)^2; \R^2))$ for any $r>0$ is the velocity field, $w = \nabla \times u$ is the vorticity, $w_0 \in L^2_{\text{per}}((0,1)^2;\R)$ is the initial vorticity,  $\nu \in \R_+$ is the viscosity coefficient, and $f \in L^2_{\text{per}}((0,1)^2;\R)$ is the forcing function. We are interested in learning the operator mapping the vorticity up to time 10 to the vorticity up to some later time $T > 10$,
$\Ftrue: C([0,10]; H^r_{\text{per}}((0,1)^2; \R)) \to C((10,T]; H^r_{\text{per}}((0,1)^2; \R))$
defined by $w|_{(0,1)^2 \times [0,10]} \mapsto w|_{(0,1)^2 \times (10,T]}$. 
Given the vorticity it is easy to derive the velocity. While vorticity is harder to model compared to velocity, it provides more information. By formulating the problem on vorticity, the neural network models mimic the pseudo-spectral method. 
We experiment with the viscosities $\nu = 1\mathrm{e}{-3}, 1\mathrm{e}{-4}, 1\mathrm{e}{-5}$, decreasing the final time $T$ as the dynamic becomes chaotic.
Since the baseline methods are not resolution-invariant, we fix the resolution to be $64 \times 64$ for both training and testing.

%The 2d Navier Stokes equation is a non-linear, time-dependent equation. We consider the map of vorticity $w = \nabla \times u$ on the domain $D \times [0,T]$ from the first $[0,10]$ time interval to the full time interval $[10,T]$:  $\Ftrue: w|_{D \times [0,10]} \mapsto w|_{D \times [10,T]}$. We test with viscosity $\nu = 1\mathrm{e}{-3}, 1\mathrm{e}{-4}, 1\mathrm{e}{-5}$. For smaller viscosity we use smaller time intervals to balance the difficulty.

\begin{table}[h]
\caption{Benchmarks on Navier Stokes (fixing resolution $64 \times 64$ for both training and testing)}
\label{table:ns}
\begin{center}
\begin{tabular}{l|rc|cccc}
\multicolumn{1}{c}{} 
&\multicolumn{1}{c}{}
&\multicolumn{1}{c}{} 
&\multicolumn{1}{c}{} 
&\multicolumn{1}{c}{}\\
 & {\bf Parameters}& {\bf Time}& $\nu=1\mathrm{e}{-3}$ &$\nu=1\mathrm{e}{-4}$ &$\nu=1\mathrm{e}{-4}$ & $\nu=1\mathrm{e}{-5}$\\
 {\bf Config}&& {\bf per} &$T=50$ &$T=30$ &$T=30$ & $T=20$\\
 && {\bf epoch} &$ N=1000$ &$ N=1000$ &$N=10000$ & $ N=1000$\\
\hline 
FNO-3D    & $6,558,537$ & $38.99s$ &${\bf 0.0086}$ &$0.1918$ &${\bf 0.0820}$  &$0.1893$ \\
FNO-2D    & $414,517$ & $127.80s$ &$0.0128 $ &${\bf 0.1559}$ &$0.0834$  &${\bf 0.1556}$ \\
% \hline
U-Net       & $24,950,491$ & $48.67s$ &$0.0245 $ &$0.2051$ &$0.1190$  &$0.1982$ \\
TF-Net       & $7,451,724$ & $47.21s$ &$0.0225 $ &$0.2253$ &$0.1168$  &$0.2268$ \\
ResNet     &$266,641$ & $78.47s$ &$0.0701 $ &$0.2871$ &$0.2311$  &$0.2753$ \\
\hline 
\end{tabular}
\end{center}
% {\small The first column is the number of parameters in the network; the second column is the training time per epoch.}
\end{table}

As shown in Table \ref{table:ns}, the FNO-3D has the best performance when there is sufficient data ($\nu=1\mathrm{e}{-3}, N=1000$ and $\nu=1\mathrm{e}{-4}, N=10000$). For the configurations where the amount of data is insufficient ($\nu=1\mathrm{e}{-4}, N=1000$ and $\nu=1\mathrm{e}{-5}, N=1000$), all methods have $>15\%$ error with FNO-2D achieving the lowest. Note that we only present results for spatial resolution $64 \times 64$ since all benchmarks we compare against are designed for this resolution. Increasing it degrades their performance while FNO achieves the same errors.  

\paragraph{2D and 3D Convolutions.}
FNO-2D, U-Net, TF-Net, and ResNet all do 2D-convolution in the spatial domain and recurrently propagate in the time domain (2D+RNN). The operator maps the solution at the previous $10$ time steps to the next time step (2D functions to 2D functions). On the other hand, FNO-3D performs convolution in space-time. It maps the initial time steps directly to the full trajectory (3D functions to 3D functions).
The 2D+RNN structure can propagate the solution to any arbitrary time $T$ in increments of a fixed interval length $\Delta t$, while the Conv3D structure is fixed to the interval $[0, T]$ but can transfer the solution to an arbitrary time-discretization. We find the 3-d  method to be more expressive and easier to train compared to its RNN-structured counterpart.  

%In general the 3D method is more expressive compared to the 2D method; it is also faster and easier to train compared to the RNN structure.

\subsection{Zero-shot super-resolution.}
\label{sec:superresolution}
The neural operator is mesh-invariant, so it can be trained on a lower resolution and evaluated at a higher resolution, without seeing any higher resolution data (zero-shot super-resolution).
Figure \ref{fig:1} shows an example where we train the FNO-3D model on $64 \times 64 \times 20$ resolution data in the setting above with ($\nu=1\mathrm{e}{-4}, N=10000$) and transfer to $256 \times 256 \times 80$ resolution, demonstrating super-resolution in space-time. Fourier neural operator is the only model among the benchmarks (FNO-2D, U-Net, TF-Net, and ResNet) that can do zero-shot super-resolution. And surprisingly, it can do super-resolution not only in the spatial domain but also in the temporal domain.

\subsection{Bayesian Inverse Problem}
\label{sec:bayesian}
In this experiment, we use a function space Markov chain Monte Carlo (MCMC) method \citep{Cotter_2013} to draw samples from the posterior distribution of the initial vorticity in Navier-Stokes given sparse, noisy observations at time $T=50$. We compare the Fourier neural operator acting as a surrogate model with the traditional solvers used to generate our train-test data (both run on GPU). We generate 25,000 samples from the posterior (with a 5,000 sample burn-in period), requiring 30,000 evaluations of the forward operator.
%Given a prior, we run \zongyi{number A} MCMC steps and evaluate \zongyi{number B} instances of the forward map with Fourier neural operator and the traditional solver respectively. 

As shown in Figure \ref{fig:baysian} (Appendix \ref{app:bayesian}), FNO and the traditional solver recover almost the same posterior mean which, when pushed forward, recovers well the late-time dynamic of Navier Stokes. 
In sharp contrast, FNO takes $0.005s$ to evaluate a single instance while the traditional solver, after being optimized to use the largest possible internal time-step which does not lead to blow-up, takes $2.2s$. This amounts to $2.5$ minutes for the MCMC using FNO and over $18$ hours for the traditional solver. Even if we account for data generation and training time (offline steps) which take $12$ hours, using FNO is still faster! Once trained, FNO can be used to quickly perform multiple MCMC runs for different initial conditions and observations, while the traditional solver will take $18$ hours for every instance. Furthermore, since FNO is differentiable, it can easily be applied to PDE-constrained optimization problems without the need for the adjoint method.

%It is because data-driven methods such as FNO do not require roll-out in time; it has $4$ Fourier layers in total, while the traditional solvers need to have a time step $\Delta t = 0.01$ to maintain a reasonable accuracy. Other reasons could attribute to better parallelization of deep learning methods on GPU. And we acknowledge there could be more efficient implementation of the traditional solvers. The inverse problems of Burgers and Darcy have a similar result \zongyi{a figure in the appendix}. 

\paragraph{Spectral analysis.}
Due to the way we parameterize $R_\phi$, the function output by (\ref{eq:Fourier}) has at most $k_{\text{max},j}$ Fourier modes per channel. This, however, does not mean that the Fourier neural operator can only approximate functions up to $k_{\text{max},j}$ modes. Indeed, the activation functions which occur between integral operators and the final decoder network $Q$ recover the high frequency modes.
As an example, consider a solution to the Navier-Stokes equation with viscosity $\nu=1\mathrm{e}{-3}$. Truncating this function at $20$ Fourier modes yields an error around $2\%$ 
% as shown in Figure \ref{fig:spectral2} (Appendix \ref{app:sepctral}), 
while our Fourier neural operator learns the parametric dependence and produces approximations to an error of $\leq 1\%$ with only $k_{\text{max},j}=12$ parameterized modes.
%The performance of the Fourier operator ties to the spectrum decay of the equation. As shown in Figure \ref{fig:spectral} (Appendix), all the equations we test on have exponential decays.
%But, for harder equations such as Navier-Stokes, the decay is slower: for $\nu=1\mathrm{e}{-3}$, truncated at $k_{max,i}=20$, the decomposition error is still around $2\%$. Notice our methods can achieve $\leq 1\%$ error at $k_{max,i}=12$, even better than the spectral decomposition.
%Although the Fourier layers we used cannot recover higher modes by itself, the Fourier neural operator can still output higher modes because of the non-linear transformation $Q$ in the end.

\paragraph{Non-periodic boundary condition.} Traditional Fourier methods work only with periodic boundary conditions. However, the Fourier neural operator does not have this limitation. This is due to the linear transform $W$ (the bias term) which keeps the track of non-periodic boundary. As an example, the Darcy Flow and the time domain of Navier-Stokes have non-periodic boundary conditions, and the Fourier neural operator still learns the solution operator with excellent accuracy.

\section{Discussion and Conclusion}
% \begin{itemize}
%     \item activation function on the spatial domain
%     \item Need uniform grid for FFT, Best when resolution is a power of 2
%     \item Can work on non-periodic boundary
%     \item recurrent structure
%     \item image problem
% \end{itemize}

%\textbf{Activation functions on the spatial domain.} We choose to apply the activation functions on the spatial domain. One can instead directly apply activation functions on the Fourier domain. But in practice, it doesn't work as well, because the pointwise activation function in the Fourier domain is equivalent to spatial convolution, which loses the meaning of the activation functions. We need the activation functions on the spatial domain to recover the higher Fourier modes.
\textbf{Requirements on Data.} Data-driven methods rely on the quality and quantity of data. To learn Navier-Stokes equation with viscosity $\nu=1\mathrm{e}{-4}$, we need to generate $N=10000$ training pairs $\{a_j,u_j\}$ with the numerical solver. However, for more challenging PDEs, generating a few training samples can be already very expensive. A future direction is to combine neural operators with numerical solvers to levitate the requirements on data.
\textbf{Recurrent structure.} The neural operator has an iterative structure that can naturally be formulated as a recurrent network where all layers share the same parameters without sacrificing performance. (We did not impose this restriction in the experiments.)
\textbf{Computer vision.} Operator learning is not restricted to PDEs. Images can naturally be viewed as real-valued functions on 2-d domains and videos simply add a temporal structure.
Our approach is therefore a natural choice for problems in computer vision where invariance to discretization crucial is important \citep{chi2020fast}. 
%Real life problems such as image classification can also be formulated as operator learning.
%We observe the Fourier layers work as well as convolutional layers on MNIST and CIFAR10 when the numbers of layers are few. But Fourier layers are less compatible with pooling and the performance doesn't improve when stacking more layers. So far it cannot parallel VGG or ResNet on image classification.

\section*{Acknowledgements}
The authors want to thank Ray Wang and Rose Yu for meaningful discussions.
Z. Li gratefully acknowledges the financial support from the Kortschak Scholars Program.
A. Anandkumar is supported in part by Bren endowed chair, LwLL grants, Beyond Limits, Raytheon, Microsoft, Google, Adobe faculty fellowships, and DE Logi grant. 
K. Bhattacharya, N. B. Kovachki, B. Liu, and A. M. Stuart gratefully acknowledge the financial support of the Army Research Laboratory through the Cooperative Agreement Number W911NF-12-0022. Research was sponsored by the Army Research Laboratory and was accomplished under Cooperative Agreement Number W911NF-12-2-0022. 
The views and conclusions contained in this document are those of the authors and should not be interpreted as representing the official policies, either expressed or implied, of the Army Research Laboratory or the U.S. Government. The U.S. Government is authorized to reproduce and distribute reprints for Government purposes notwithstanding any copyright notation herein. 

\bibliography{iclr2021}
\bibliographystyle{iclr2021}

\newpage
\appendix
\section{Appendix}

\subsection{Table of notations}
A table of notations is given in Table \ref{table:notations}.

\begin{table}[h]
\caption{table of notations}
\label{table:notations}
\begin{center}
\begin{tabular}{|l|l|}
\multicolumn{1}{c}{\bf Notation} 
&\multicolumn{1}{c}{\bf Meaning}\\
\hline 
{\bf Operator learning} &\\
$D \subset \R^d$  & The spatial domain for the PDE \\
$x \in D$  & Points in the the spatial domain \\
$a \in \A = (D;\R^{d_a})$  & The input coefficient functions \\
$u \in \U = (D;\R^{d_u})$  & The target solution functions \\
$D_j$ & The discretization of $(a_j, u_j)$\\
$\Ftrue: \A \to \U$  & The operator mapping the coefficients to the solutions\\
$\mu$ & A probability measure where $a_j$ sampled from.\\
\hline
{\bf Neural operator} &\\
$v(x) \in \R^{d_v}$  & The neural network representation of $u(x)$ \\
$d_a$ & Dimension of the input $a(x)$.\\
$d_u$ & Dimension of the output $u(x)$.\\
$d_v$ & The dimension of the representation $v(x)$\\
$\kappa : \R^{2(d+1)} \to \R^{d_v \times d_v}$  & The kernel maps $(x,y,a(x),a(y))$ to a $d_v \times d_v$ matrix\\
$\phi$  & The parameters of the kernel network $\kappa$ \\
$t = 0,\ldots,T$  & The time steps (layers)  \\
$\sigma $  & The activation function  \\
\hline 
{\bf Fourier operator} &\\
$\cG, \cG^{-1}$ & Fourier transformation and its inverse.\\
$R$ & The linear transformation applied on the lower Fourier modes.\\
$W$ & The linear transformation (bias term) applied on the spatial domain. \\
$k$ & Fourier modes / wave numbers.\\
$k_{max}$ & The max Fourier modes used in the Fourier layer.\\
\hline 
{\bf Hyperparameters} &\\
$N$  & The number of training pairs. \\
$n$  & The size of the discretization. \\
$s$  & The resolution of the discretization $(s^d = n)$. \\
$\nu$ & The viscosity. \\
$T$ & The time interval $[0,T]$ for time-dependent equation. \\
\hline
\end{tabular}
\end{center}
\end{table}

\subsection{Spectral Analysis}
\label{app:sepctral}
\begin{figure}[h]
    \centering
    \includegraphics[width=4.5cm]{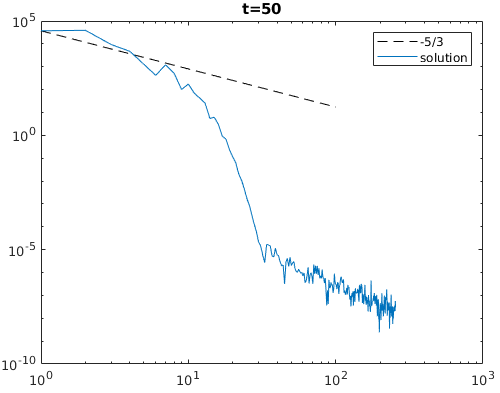}
    \includegraphics[width=4.5cm]{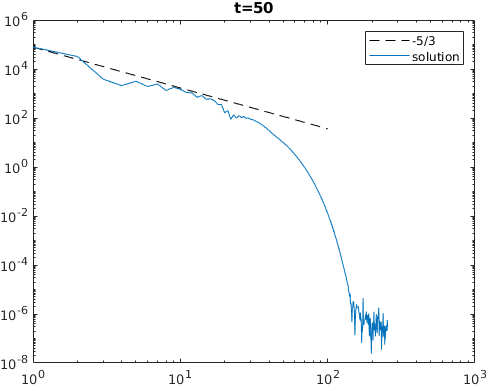}
    \includegraphics[width=4.5cm]{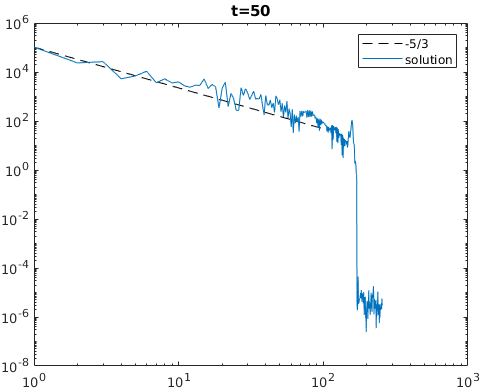}\\
    {\small The spectral decay of the Navier-stokes equation data we used in section \ref{sec:ns}. The y-axis is the spectrum; the x-axis is the wavenumber $|k| = k_1 + k_2$.  }
    \caption{Spectral Decay of Navier-Stokes equations}
    \label{fig:spectral1}
\end{figure}

The spectral decay of the Navier Stokes equation data is shown in Figure \ref{fig:spectral1}. The spectrum decay has a slope $k^{-5/3}$, matching the energy spectrum in the turbulence region. And we notice the energy spectrum does not decay along with time.

\begin{figure}[h]
    \centering
    \includegraphics[width=6cm]{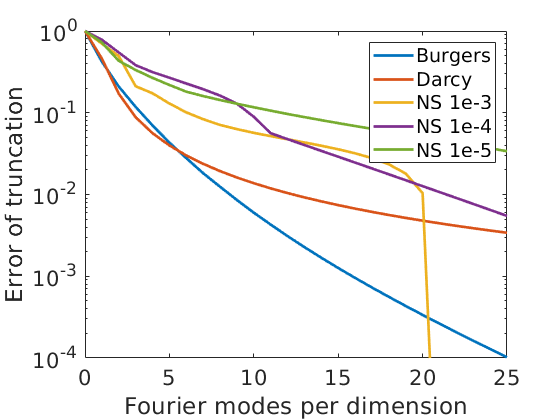}\\
    {\small The error of truncation in one single Fourier layer without applying the linear transform $R$. The y-axis is the normalized truncation error; the x-axis is the truncation mode $k_{max}$.  }
    \caption{Spectral Decay in term of $k_{max}$ }
    \label{fig:spectral2}
\end{figure}

We also present the spectral decay in term of the truncation $k_{max}$ defined in \ref{sec:fourier} as shown in Figure\ref{fig:spectral2}. We note all equations (Burgers, Darcy, and Navier-Stokes with $\nu \leq 1\mathrm{e}{-4}$ ) exhibit high frequency modes. Even we truncate at $k_{max} = 12$ in the Fourier layer, the Fourier neural operator can recover the high frequency modes.

\subsection{Data generation}
In this section, we provide the details of data generator for the three equation we used in Section \ref{sec:numerics}.
\subsubsection{Burgers Equation}
\label{app:burgers}
Recall the 1-d Burger's equation on the unit torus:
\begin{align*}
    \begin{split}
    \partial_t u(x,t) + \partial_x ( u^2(x,t)/2) &= \nu \partial_{xx} u(x,t), \qquad x \in (0,1), t \in (0,1] \\
    u(x,0) &= u_0(x), \qquad \qquad \: x \in (0,1).
    \end{split}
\end{align*}
The initial condition $ u_0(x)$ is generated according to 
\(u_0 \sim \mu\) where \(\mu = \mathcal{N}(0,625(-\Delta + 25I)^{-2})\) with periodic boundary conditions. We set the viscosity to \(\nu = 0.1\) and solve the equation using a split step method where the heat equation part is solved exactly in Fourier space then the non-linear part is advanced, again in Fourier space, using a very fine forward Euler method. We solve on a spatial mesh with resolution $2^{13} = 8192$ and use this dataset to subsample other resolutions.

\subsubsection{Darcy Flow}
\label{app:darcy}
The 2-d Darcy Flow is a second-order linear elliptic equation of the form
\begin{align*}
\begin{split}
- \nabla \cdot (a(x) \nabla u(x)) &= f(x) \qquad x \in (0,1)^2 \\
u(x) &= 0 \qquad \quad \:\: x \in \partial (0,1)^2.
\end{split}
\end{align*}
The coefficients $a(x)$ are generated according to \(a \sim \mu \) where \(\mu = \psi_{\#} \mathcal{N}(0,(-\Delta + 9I)^{-2})\)
with zero Neumann boundary conditions on the Laplacian.
The mapping \(\psi : \R \to \R\) takes the value $12$ on the positive part of the real line and $3$ on the negative and the push-forward is defined pointwise. The forcing is kept fixed $f(x) = 1$. Such constructions are prototypical models for many physical systems such as permeability in subsurface flows and material microstructures in elasticity. Solutions \(u\) are obtained by using a second-order finite difference scheme on a $421 \times 421$ grid. Different resolutions are downsampled from this dataset. 

\subsubsection{Navier-Stokes Equation}
\label{app:ns}
Recall the 2-d Navier-Stokes equation for a viscous, incompressible fluid in vorticity form on the unit torus:
% {\bf equations under construction}\\
\begin{align*}
\begin{split}
\partial_t w(x,t) + u(x,t) \cdot \nabla w(x,t) &= \nu \Delta w(x,t) + f(x), \qquad x \in (0,1)^2, t \in (0,T]  \\
\nabla \cdot u(x,t) &= 0, \qquad \qquad \qquad \qquad \quad x \in (0,1)^2, t \in [0,T]  \\
w(x,0) &= w_0(x), \qquad \qquad \qquad \quad x \in (0,1)^2.
\end{split}
\end{align*}
The initial condition $ w_0(x)$ is generated according to 
\(w_0 \sim \mu\) where \(\mu = \mathcal{N}(0,7^{3/2}(-\Delta + 49I)^{-2.5})\) with periodic boundary conditions. The forcing is kept fixed $f(x) = 0.1(\sin(2\pi(x_1+x_2)) + \cos(2\pi(x_1 + x_2)))$. The equation is solved using the stream-function formulation with a pseudospectral method. First a Poisson equation is solved in Fourier space to find the velocity field. Then the vorticity is differentiated and the non-linear term is computed is physical space after which it is dealiased. Time is advanced with a Crank–Nicolson update where the non-linear term does not enter the implicit part. 
All data are generated on a $256 \times 256$ grid and are downsampled to $64 \times 64$. We use a time-step of $1\mathrm{e}{-4}$ for the Crank–Nicolson scheme in the data-generated process where we record the solution every $t=1$ time units. The step is increased to $2\mathrm{e}{-2}$ when used in MCMC for the Bayesian inverse problem.

\subsection{Results of Burgers' equation and Darcy Flow}
The details error rate on Burgers' equation and Darcy Flow are listed in Table \ref{table:burgers} and Table \ref{table:darcy}.

\begin{table}[h]
\caption{ Benchmarks on 1-d Burgers' equation} 
\label{table:burgers}
\begin{center}
\begin{tabular}{l|llllllll}
\multicolumn{1}{c}{\bf Networks}
&\multicolumn{1}{c}{\bf $s=256$}
&\multicolumn{1}{c}{\bf $s=512$}
&\multicolumn{1}{c}{\bf $s=1024$}
&\multicolumn{1}{c}{\bf $s=2048$} 
&\multicolumn{1}{c}{\bf $s=4096$}
&\multicolumn{1}{c}{\bf $s=8192$}\\
\hline 
NN       &$0.4714$ &$0.4561$
&$0.4803$ &$0.4645$ &$0.4779$ &$0.4452$ \\
GCN           &$0.3999$ &$0.4138$
&$0.4176$   &$0.4157$  &$0.4191$ &$0.4198$\\
FCN         &$0.0958$ &$0.1407$
&$0.1877$   &$0.2313$  &$0.2855$ &$0.3238$\\
PCANN       &$0.0398$ &$0.0395$
&$0.0391$   &$0.0383$  &$0.0392$ &$0.0393$\\
\hline 
GNO     &$0.0555$ &$0.0594$ &$0.0651$   &$0.0663$  &$0.0666$ &$0.0699$\\
LNO      &$0.0212$ &$0.0221$
   &$0.0217$  &$0.0219$ &$0.0200$ &$0.0189$\\
MGNO      &$0.0243$ &$0.0355$
   &$0.0374$  &$0.0360$ &$0.0364$ &$0.0364$\\
FNO     &$\textbf{0.0149}$ &$\textbf{0.0158}$
   &$\textbf{0.0160}$  &$\textbf{0.0146}$ &$\textbf{0.0142}$ &$\textbf{0.0139}$\\
\hline 
\end{tabular}
\end{center}
\end{table}

\begin{table}[h]
\caption{Benchmarks on 2-d Darcy Flow}
\label{table:darcy}
\begin{center}
\begin{tabular}{l|llll}
\multicolumn{1}{c}{\bf Networks} 
&\multicolumn{1}{c}{\bf $s=85$}
&\multicolumn{1}{c}{\bf $s=141$} 
&\multicolumn{1}{c}{\bf $s=211$}
&\multicolumn{1}{c}{\bf $s=421$}\\
\hline
NN       &$0.1716$  &$0.1716$  &$0.1716$ &$0.1716$\\
FCN       &$0.0253$  &$0.0493$  &$0.0727$ & $0.1097$\\
PCANN      &$0.0299$  &$0.0298$  &$0.0298$ & $0.0299$\\
RBM    &$0.0244$ &$0.0251$ &$0.0255$ &$0.0259$ \\
\hline 
GNO     &$0.0346$   &$0.0332$  &$0.0342$ &$0.0369$\\
LNO     &$0.0520$  &$0.0461$  &$0.0445$ &$-$\\
MGNO     &$0.0416$   &$0.0428$  &$0.0428$ &$0.0420$\\
FNO     &$\textbf{0.0108}$  &$\textbf{0.0109}$  &$\textbf{0.0109}$ &$\textbf{0.0098}$\\
\hline 
\end{tabular}
\end{center}
\end{table}

\subsection{Bayesian Inverse Problem}
\label{app:bayesian}
\begin{figure}[t]
    \centering
    \begin{subfigure}[b]{0.32\textwidth}
        \includegraphics[width=\textwidth]{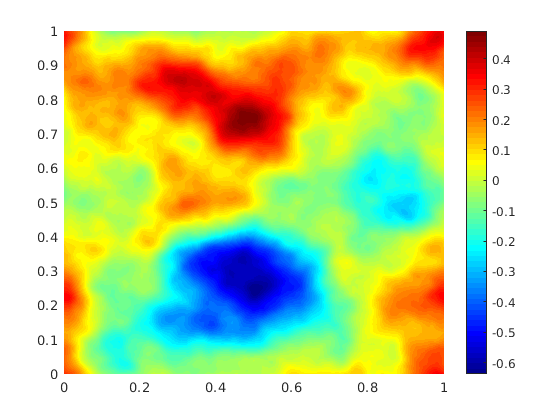}
        %\caption{\(\mug\)}
    \end{subfigure}
    \begin{subfigure}[b]{0.32\textwidth}
        \includegraphics[width=\textwidth]{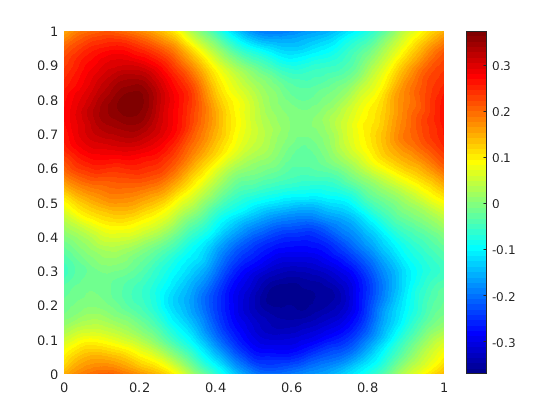}
        %\caption{\(\mul\)}
    \end{subfigure}
    \begin{subfigure}[b]{0.32\textwidth}
        \includegraphics[width=\textwidth]{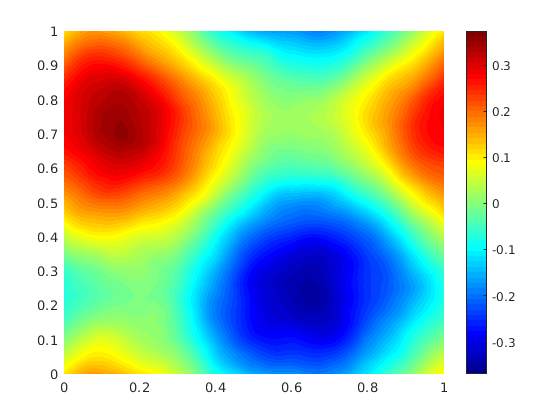}
        %\caption{\(\mup\)}
    \end{subfigure}
    
    \begin{subfigure}[b]{0.32\textwidth}
        \includegraphics[width=\textwidth]{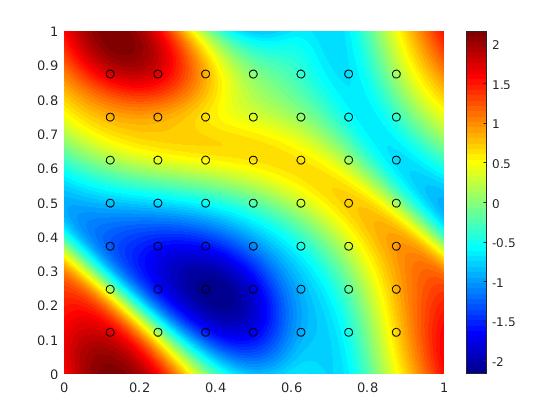}
        %\caption{\(\mug\)}
    \end{subfigure}
    \begin{subfigure}[b]{0.32\textwidth}
        \includegraphics[width=\textwidth]{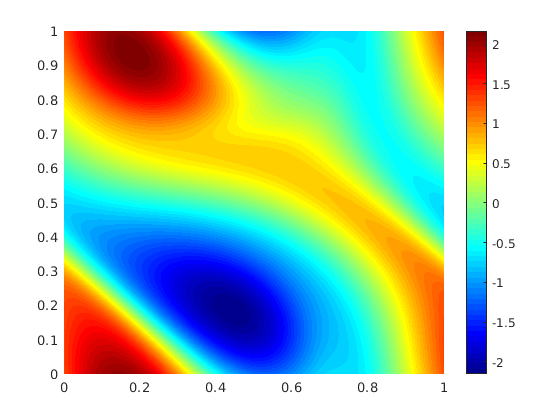}
        %\caption{\(\mul\)}
    \end{subfigure}
    \begin{subfigure}[b]{0.32\textwidth}
        \includegraphics[width=\textwidth]{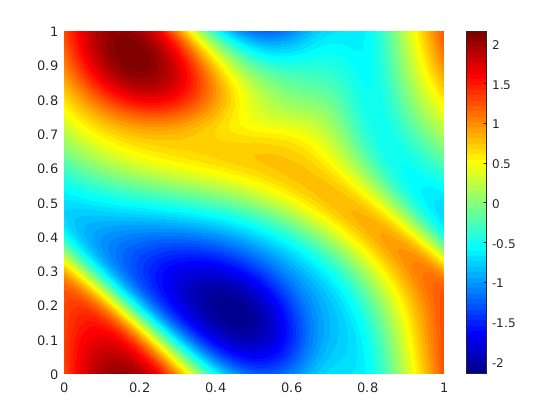}
        %\caption{\(\mup\)}
    \end{subfigure}
{\small The top left panel shows the true initial vorticity while bottom left panel shows the true observed vorticity at $T=50$ with black dots indicating the locations of the observation points placed on a $7 \times 7$ grid. The top middle panel shows the posterior mean of the initial vorticity given the noisy observations estimated with MCMC using the traditional solver, while the top right panel shows the same thing but using FNO as a surrogate model. The bottom middle and right panels show the vorticity at $T=50$ when the respective approximate posterior means are used as initial conditions. }
    \caption{Results of the Bayesian inverse problem for the Navier-Stokes equation. } 
    \label{fig:baysian} 
\end{figure}

Results of the Bayesian inverse problem for the Navier-Stokes equation are shown in Figure \ref{fig:baysian}. It can be seen that the result using Fourier neural operator as a surrogate is as good as the result of the traditional solver.

\end{document}